\pdfoutput=1

\documentclass[11pt]{article}

\usepackage[]{acl}

\usepackage{times}
\usepackage{latexsym}

\usepackage[T1]{fontenc}

\usepackage[utf8]{inputenc}

\usepackage{microtype}
\usepackage{url}            
\usepackage{booktabs}       
\usepackage{amsfonts}       
\usepackage{nicefrac}       
\usepackage{cuted}
\usepackage{enumitem}
\usepackage{bm}
\usepackage{rotating}
\usepackage{multirow}
\usepackage[percent]{overpic}

\usepackage{listings} 

\usepackage{tabulary}
\usepackage{tabularx}
\usepackage{fancyvrb}
\usepackage{fvextra}

\usepackage{longtable}
\usepackage{graphicx}
\graphicspath{ {./img/} }

\usepackage{inconsolata}

\usepackage[disable,textsize=tiny]{todonotes}
\definecolor{forestgreen(web)}{rgb}{0.13, 0.55, 0.13}

\title{Improving Generalization in Semantic Parsing \\
by Increasing Natural Language Variation
}

\author{Irina Saparina  \and Mirella Lapata \\  
  Institute for Language, Cognition and Computation \\
  School of Informatics, University of Edinburgh  \\
  10 Crichton Street, Edinburgh EH8 9AB\\
    \texttt{i.saparina@sms.ed.ac.uk} \qquad  \texttt{mlap@inf.ed.ac.uk}}

\begin{document}
\maketitle
\begin{abstract}
Text-to-SQL semantic parsing has made significant progress in recent years, with various models  demonstrating impressive performance on the challenging Spider benchmark. However, it has also been shown that these models often struggle to generalize even when faced with small perturbations of previously (accurately) parsed expressions. 
This is mainly due to the linguistic form of questions in  Spider which are overly specific,  unnatural, and display limited variation. 
In this work, we use data augmentation to enhance the robustness of text-to-SQL parsers against natural language variations. Existing approaches generate question reformulations either via models trained on Spider or only introduce local changes. 
In contrast,  we leverage the capabilities of large language models to generate more realistic and diverse questions.
Using only a few prompts, we achieve a two-fold increase in the number of questions in Spider. Training on this augmented 
 dataset yields substantial improvements on a range of evaluation sets, including robustness benchmarks and out-of-domain data.\footnote{Model checkpoints and data are available at \href{https://github.com/saparina/Text2SQL-NLVariation}{\nolinkurl{github.com/saparina/Text2SQL-NLVariation}}}
\end{abstract}

\section{Introduction}
\label{sec:Introduction}
Semantic parsing is the task of mapping natural language utterances to
machine-interpretable expressions such as SQL queries or logical
forms. It has emerged as an important component in
many natural language interfaces \citep{Ozcan:ea:2020} with
applications in robotics~\citep{dukes-2014-semeval}, question
answering~\citep{zhongSeq2SQL2017,yu-etal-2018-spider}, dialogue
systems \cite{artzi-zettlemoyer-2011-bootstrapping}, and the Internet
of Things~\citep{10.1145/3038912.3052562}.

The release of the Spider dataset \citep{yu-etal-2018-spider} marked
an important milestone in text-to-SQL semantic parsing. Apart from its
considerable size, Spider stands out for including complex and nested
queries, and databases from various domains. Importantly, it
exemplifies a cross-domain generalization setting, i.e.,~models
trained on Spider are expected to parse natural language questions for any
given database, even in previously unseen domains.  In practice,
models trained on Spider degrade significantly when tested on
different databases from \emph{other} datasets, for example, on
real-world data from Kaggle and Stack Exchange websites
\citep{suhr-etal-2020-exploring, lee-etal-2021-kaggledbqa, hazoom-etal-2021-text}.

The linguistic composition of questions in Spider contributes to this
performance gap.  Unlike real-world applications where user questions
may be concise, ambiguous, and necessitate commonsense reasoning or
domain-specific knowledge, questions in Spider are often overly
explicit, directly mentioning database entities even when such
information is unnecessary for inferring the underlying intent. An
example is shown in Figure~\ref{tab:example}, the first question includes
redundant details (e.g., \textsl{customer}, \textsl{first name},  \textsl{last name}) which serve as
references to databases entities.  Omitting these details would not
change the meaning of the question but rather make it
more colloquial. Due to the limited diversity of questions, Spider falls short in providing enough
examples for learning essential skills such as grounding and reasoning. 
As a result, models tend to overfit to Spider-style questions, and
even minor perturbations in how questions are phrased  lead to 
considerable performance decrease, sometimes up to~22\% \citep{gan-etal-2021-exploring, deng-etal-2021-structure,
  pi-etal-2022-towards, chang2023drspider}.

Efforts to automatically increase its diversity often rely on text
generation models trained on the same Spider data and unavoidably
inherit its characteristics \cite{Zhong-etal-2020-grounded, wang-etal-2021-learning-synthesize, wu-etal-2021-data, Jiang2022}. 
In this work, we propose to augment the training data for text-to-SQL parsers with more realistic and diverse question reformulations. We leverage the capabilities of large language models for rewriting utterances and devise prompts designed to enhance model robustness against linguistic variations. Our prompts consist solely of instructions and questions and are easy to use.
We train three state-of-the-art {text-to-SQL} parsers on Spider \cite{yu-etal-2018-spider}  with augmentations generated by our approach. Extensive experiments show that a \mbox{two-fold} increase in the number of questions substantially improves model generalization ability. Our augmentations increase  \emph{robustness} against question perturbations when models are evaluated on the challenging Dr.Spider sets \cite{chang2023drspider} and deliver improvements in a \emph{zero-shot} setting, when models are tested on out-of-domain datasets like GeoQuery \cite{zelle1996learning} and KaggleDBQA \cite{lee-etal-2021-kaggledbqa}.  

Our contributions are three-fold: a proposal of rewrite operations to render questions more diverse and natural; a methodology for augmenting existing datasets based on the proposed reformulations; and empirical results  validating our approach improves generalization across models and datasets. 
 
\section{Related Work}

\paragraph{Out-of-domain Generalization} 
Several
datasets have been released  to facilitate the
development of models with generalization capabilities. WikiSQL
\cite{zhongSeq2SQL2017} is a \mbox{large-scale} benchmark with different
databases  but only one table. As a result,
WikiSQL queries are relatively easy to parse due to the use of a
limited set of operations.  Spider \citep{yu-etal-2018-spider},
contains multiple tables per database which result in complex 
SQL queries.

\citet{suhr-etal-2020-exploring} examine the performance of
Spider-trained models on  datasets varying in terms of the questions being asked, the database structure, and
SQL style. They discover that a key challenge in achieving
generalization lies in linguistic variation,
and propose augmenting Spider's training set with WikiSQL data.  Our
work addresses the problem of question diversity in Spider, without
compromising its complex query structures or multi-table database nature.
We evaluate our approach on GeoQuery \citep{zelle1996learning},
a dataset similar to Spider in terms of
database structure and SQL queries but different in the style of
questions. We also report results on KaggleDBQA
\cite{lee-etal-2021-kaggledbqa}, a dataset with real-world
databases and questions created by
users with access to field descriptions rather than database
schemas.

\textsc{Bird} \cite{li2023llm}  is a recently released a text-to-SQL benchmark, aiming to highlight real-world challenges with large-scale databases which often contain  dirty and noisy values and how to express SQL queries to improve execution speed. They show that incorporating manually annotated external knowledge that includes synonyms improves performance. Our augmentations can be viewed as an alternative to this approach; we \emph{learn} language variations \emph{without} oracle knowledge. 

\paragraph{Robustness to Perturbations} Another challenge for 
text-to-SQL parsers is robustness to small perturbations. Previous
studies evaluate robustness in the single-domain setting
\citep{Huang-etal-2021-robustness} and across databases, e.g.,~by
removing or paraphrasing explicit mentions of database entities
(Spider-Realistic; \citealt{deng-etal-2021-structure}) or by
substituting such mentions with synonyms (\mbox{Spider-Syn};
\citealt{gan-etal-2021-towards}).  Other work explores the effect of
perturbations in the database schema \cite{pi-etal-2022-towards} and
also in questions \cite{ma_and_wang2021}. Recently,
\citet{chang2023drspider} released Dr.Spider, a comprehensive
robustness benchmark with a wide range of perturbations in
the database schema, questions, and SQL semantics. We evaluate our
approach on their ``question sets'' which cover a broader range of
language variations compared to previous efforts.

\paragraph{Data Augmentation} Several data augmentation and
adversarial training techniques have been proposed to support SQL
queries executed on a single table
\cite{Li:ea:2019,Radhakrishnan-etal-2020-COLLOQL} and multiple tables
\cite{Zhong-etal-2020-grounded, 
wang-etal-2021-learning-synthesize,  wu-etal-2021-data, deng-etal-2021-structure,wu-etal-2021-data, Jiang2022}. 
Augmentations in earlier work \cite{gan-etal-2021-towards, deng-etal-2021-structure,ma_and_wang2021,
  Huang-etal-2021-robustness}
 target specific linguistic expressions like synonyms 
or paraphrases. 
We
leverage the capabilities of (very) large languages models (LLMs;
\citealt{brown2020language,chowdhery2022palm}) to generate linguistically diverse
 natural language questions. Recent efforts
\cite{dai2023auggpt,he2023annollm} have shown that LLMs can serve as
annotators when given sufficient guidance and examples mainly for text
classification tasks.

\section{Motivation}
\subsection{Problem Formulation} 
Semantic parsing aims to translate a natural language
utterance into a formal representation of its meaning. 
We  focus on meaning representations in the form of SQL
queries that can be executed in some  database to retrieve an answer or denotation.
In the cross-domain setting, the parser is not limited to a specific
database and can be in theory applied to arbitrary databases and
questions.  In practice, this task is more or less complex
depending on the database in hand, i.e.,~the number of tables and
values, the naming conventions used for tables and columns, the way
values are formatted, and specific domain characteristics. We
do not consider these challenges in this work, focusing instead on
generalization issues that arise from the variation of questions in
natural language.

\subsection{Types of Utterances in Semantic Parsing} Recent work has demonstrated the importance of wording in semantic parsing, indicating that certain question formulations can be more difficult to parse than others \citep{Radhakrishnan-etal-2020-COLLOQL, gan-etal-2021-towards, deng-etal-2021-structure, chang2023drspider}. 

The level of difficulty for a question can be influenced by the amount of task-specific
background knowledge used to formulate it.  For instance, users familiar
with SQL and the underlying database will have some idea of the
desired program, and will be able to articulate their intentions more
precisely, e.g., by providing explicit instructions.  In contrast,
users unfamiliar with the task are more likely to ask
general questions in a colloquial style.
Figure~\ref{fig:example} illustrates different question formulations
with the same intent. The first question could have been posed by a
user who is well-versed in SQL and has knowledge of the database; it
mentions specific database entities and operations like summation and
filtering, unlike the second question which does not have any such
details.
More formally, we distinguish between two types of utterances:

\paragraph{Utterances which demonstrate prior knowledge}  are closely aligned with the
    desired programs,  highlight  logical structure
    operations,  and explicit references to database entities. 
    Such utterances resemble instructions, suggesting the user has some understanding of the desired
    program.  In  Figure~\ref{fig:example}, the
    first question falls under this category, presupposing knowledge
    of summation and filtering operations and the names of entities
    (e.g.,~\texttt{first\_name}, \texttt{last\_name})  used in the target SQL query.

\paragraph{Utterances which do not demonstrate prior
      knowledge} \hspace*{-.2cm}are general descriptions of intent, expressed in
a simple, colloquial language.  They do not provide intentional hints about the desired program, but are often ambiguous, requiring additional reasoning based on domain or common sense knowledge. In the examples shown in
Figure~\ref{fig:example}, the second question belongs to this
category, it is laconic, underspecified, and inherently natural. 

\begin{figure}[t]
\ttfamily
  \resizebox{\columnwidth}{!}{%
\begin{tabular}{|c|c|c|c|c|c|}
  \multicolumn{6}{l}{\textbf{Database: driving\_school}} \\ \hline
  \multicolumn{6}{|c|}{\textbf{Customers}} \\ \hline
  customer\_id & $\dots$ & first\_name & last\_name & $\dots$ &
  email\_address \\ \hline 
  \multicolumn{6}{c}{} \\ \hline
    \multicolumn{6}{|c|}{\textbf{Lessons}} \\ \hline
 lesson\_id & $\dots$ & customer\_id & lesson\_time & $\dots$ &
 price \\\hline
\end{tabular}
  }
   \resizebox{\columnwidth}{!}{%
     \begin{tabular}{@{}l@{~}p{7.5cm}ccc}
       \multicolumn{4}{c}{} \\ 
\multicolumn{1}{c}{}& \multicolumn{1}{c}{}& \multicolumn{1}{c}{}& \multicolumn{2}{c}{\texttt{\textbf{Prior}}} \\ 
 \multicolumn{2}{c}{\textbf{Questions}}  &  & \textbf{SQL} & \textbf{DB}\\
\textnormal{1.}&     \textnormal{Calculate the total sum of lesson times filtering the
results by selecting the customer with the first name
"Rylan" and the last name "Goodwin".} &  & \checkmark & 
  \checkmark\\
& & \multicolumn{1}{c}{}& \multicolumn{1}{c}{} & \multicolumn{1}{c}{}\\
\textnormal{2.}& \textnormal{How long did Rylan Goodwin's lesson last?}&  &  \textnormal{X} &
 \textnormal{X}\\& & \multicolumn{1}{c}{} & & \multicolumn{1}{c}{}\\
\textnormal{3.} & \textnormal{How long is the total lesson time taken by a customer
with a first name as Rylan and a last name as Goodwin?} &  &
 \textnormal{X} &  \checkmark \\
   \end{tabular}}
   \resizebox{\columnwidth}{!}{%
     \begin{tabular}{p{10.7cm}}
       \multicolumn{1}{c}{} \\ \hline
\multicolumn{1}{c}{\textbf{SQL Query}}  \\ \hline
SELECT sum(T1.lesson\_time) FROM Lessons AS T1
JOIN Customers AS T2 ON T1.customer\_id = T2.customer\_id
WHERE T2.first\_name = "Rylan" AND T2.last\_name = "Goodwin". \\ \hline
   \end{tabular}}
\caption{Different types of questions that are related to
  the same database (only relevant tables and columns are shown) and
  map to the same SQL query.}
\label{fig:example}
\end{figure}

These types of utterances  represent two important edge cases but do not cover
all possibilities.  In the context of text-to-SQL semantic parsing,
information about the database schema and its contents can also be useful
when formulating questions.  We thus introduce a third category that
falls between having  task-specific knowledge and none at all.

\paragraph{Utterances which demonstrate knowledge of  the database schema}
\hspace*{-.2cm}are general descriptions of intent but with explicit references
to related database entities. This category differs from the previous
two in the type of prior knowledge used; users are familiar with the
database schema and possibly database content but have no expertise in
query construction. The third question in Figure~\ref{fig:example}
includes explicit references to the database table (e.g.,
\texttt{\footnotesize customers}) and its columns
(e.g.,~\texttt{\footnotesize lesson\_time}, \texttt{\footnotesize
  first\_name}, \texttt{\footnotesize last\_name}). Because of that,
questions may be less coherent and natural.  In our example, the
question contains redundant details such as \textsl{first name},
\textsl{last name}, and \textsl{customer}.

Questions in Spider \citep{yu-etal-2018-spider} often include
  explicit mentions of database elements
  \citep{deng-etal-2021-structure}.  This is a by-product of Spider's 
 creation process which encouraged annotators familiar with SQL
  to formulate the questions more clearly and explicitly. In contrast,
other datasets like GeoQuery \citep{zelle1996learning} or cross-domain
KaggleDBQA \citep{lee-etal-2021-kaggledbqa} contain less explicit
questions with a smaller percentage of database entity
mentions. 
In this work, we automatically augment Spider's training set with  more general
and natural questions aiming to develop semantic parsing models that can
effectively handle all types of utterances mentioned above. 

\section{Data Generation}\label{sec:data_generation}

We augment the training set of Spider  \citep{yu-etal-2018-spider}    by leveraging large language models.  Specifically, we exploit ChatGPT's\footnote{\href{https://chat.openai.com}{chat.openai.com}} text generation capabilities  (\texttt{gpt-3.5-turbo-0301}) 
and ask it to rephrase Spider questions
(no SQL- or database-specific information is provided; see Table~\ref{tab:example}), using three  types of rewrite operations:

\begin{enumerate}[noitemsep,topsep=1mm, leftmargin=3mm] 
\item \textbf{Deletion} of words or phrases which are redundant for understanding the question's intent. For this purpose, we use two instructions: the first one  \textit{simplifies} the  question, while  the second one explicitly \textit{hides unnecessary details} that do not change the meaning. The first instruction affords ChatGPT more freedom  in rewriting the question. 
 In Table~\ref{tab:example}, examples 1--2 show how Spider questions are reformulated with these instructions.

\item \textbf{Substitution} of  words or phrases with simpler ones. We instruct ChatGPT to replace  words with their \emph{synonyms} and  also to more generally attempt to  \textit{simplify by substituting a few words} in the  question. In Table~\ref{tab:example}, examples~3--4 show how questions are rewritten  with  these instructions.

\item \textbf{Rewriting} of the entire question. Some questions can have the same meaning, despite being significantly dissimilar in their surface realisation. For example, the  questions \textsl{Where do most people live?} and \textsl{Which cities have the largest population?} are related to the same database about cities and express the same intent but have no words in common. We instruct ChatGPT to provide \textit{different ways of expressing} a question. We empirically find that ChatGPT can be too conservative at times  and also include \textit{a prompt with examples} to encourage more drastic reformulations. In Table~\ref{tab:example}, questions~5--6 show example outputs for these instructions.
\end{enumerate}

\begin{table}[t!]
\centering
\footnotesize
 \vspace{1mm}
\newcounter{Number}
\newcommand{\Item}{\stepcounter{Number}\theNumber.~}
\settowidth\tymin{\textbf{6. Instruction:}}
\begin{tabulary}{\linewidth}{@{}L@{\;}L@{}}\setcounter{Number}{0}
\textbf{\Item Instruction:} & \textbf{\texttt{Simplify}} \\ \addlinespace[3pt]
Original Question & Which club has the most female students as their members? Give me the name of the club. 
\\ \addlinespace[3pt]
Output &  Which club has the most female students?
 \\ \midrule
\textbf{\Item Instruction:} & \textbf{\texttt{Simplify by hiding details}} \\ \addlinespace[3pt]
Original Question & What is the title and credits of the course that is taught in the largest classroom (with the highest capacity)? \\ \addlinespace[3pt]
 Output &  What course is taught in the biggest classroom and what are its credits? \\ \midrule
\textbf{\Item Instruction:} & \textbf{\texttt{Simplify using synonyms}}   \\ \addlinespace[3pt]
 Original Question & What is the average duration in milliseconds of tracks that belong to Latin or Pop genre?
 \\ \addlinespace[3pt]
 Output & What is the mean length in milliseconds of Latin or Pop songs?
 \\ \midrule
\textbf{\Item Instruction:} & \textbf{\texttt{Simplify using substitutions}} \\ \addlinespace[3pt]
 Original Question & What are the locations that have gas stations owned by a company with a market value greater than 100? 
 \\ \addlinespace[3pt]
 Output &  Where are the gas stations owned by a company worth more than 100? \\ \midrule
 \textbf{\Item Instruction:} &   \textbf{\texttt{Express in a different way}} \\ \addlinespace[3pt]
  Original Question & What is the number of routes operated by the airline American Airlines whose destinations are in Italy? 
  \\ \addlinespace[3pt]
 Output &  How many routes does American Airlines have that fly to Italy?
 \\ \midrule
   \textbf{\Item Instruction:} & \textbf{\texttt{Examples of the question simplification: <$\dots$>}
   }\\ \addlinespace[3pt]
   Original Question & What are the names of body builders whose total score is higher than 300? \\ \addlinespace[3pt]
 Output &  Who are the body builders with a score over 300?\\ \midrule
   \textbf{\Item Instruction:} & \textbf{\texttt{Paraphrase}} \\ \addlinespace[3pt]
  Original Question &  Return the categories of music festivals that have the result "Awarded". \\ \addlinespace[3pt]
    Output &  List the categories of music festivals that have been recognized with awards. \\ \bottomrule
\end{tabulary}
 \caption{
Different types of augmentation generated for Spider questions. The full version of the instructions for generation is shown in Appendix~\ref{app:data_generation}, Table~\ref{tab:full_prompts}.
}
\label{tab:example}
\end{table}

We also ask ChatGPT to \textit{paraphrase}  questions (see example~7 in Table~\ref{tab:example}). This instruction may be 
viewed as a generalization of previous reformulations, however, in practice it is only somewhat helpful. ChatGPT often generates very similar versions of the original question, retaining the same details, style and structure following this instruction. 
We consider this  conservative paraphrasing strategy to be  an advantage as almost all machine-generated questions preserve the meaning of the original question. 
To verify the quality of generated paraphrases, we compute the cosine similarity between the original and generated questions.

\section{Experimental Setup}
Our experiments aim to evaluate the performance of models trained specifically for cross-database text-to-SQL parsing. We are interested in two types of generalization: robustness to controllable perturbations in utterances and adaptation to new domains with different question styles. 
Perturbations allow us to study more closely the impact of language variations, while new domains provide a more realistic and challenging setting. We first describe the datasets we use for training and evaluation and then briefly discuss the semantic parsing models and metrics we employ in our experiments. 

\begin{table}[t!]
\centering
\footnotesize
\begin{tabular}{lr}
\toprule
\multicolumn{1}{c}{Augmentation Type} & \multicolumn{1}{c}{\# examples} \\ \midrule
Simplify & 774
\\\addlinespace[3pt]
Simplify by hiding details & 1,136 
\\\addlinespace[3pt]
Simplify using synonyms & 1,285
\\\addlinespace[3pt]
Simplify using substitutions & 1,316
\\\addlinespace[3pt]
Paraphrase & 1,130
\\\addlinespace[3pt]
Express in a different way & 1,065
\\\addlinespace[3pt]
Prompt with examples &  1,256
\\\midrule
Total & 7,962
\\\bottomrule
\end{tabular}
\caption{Question reformulations generated  for Spider; number of generations per instruction.}
\label{tab:statistics}
\end{table}

\subsection{Training Datasets} 
\begin{figure}[t]
\includegraphics[width=0.93\columnwidth]{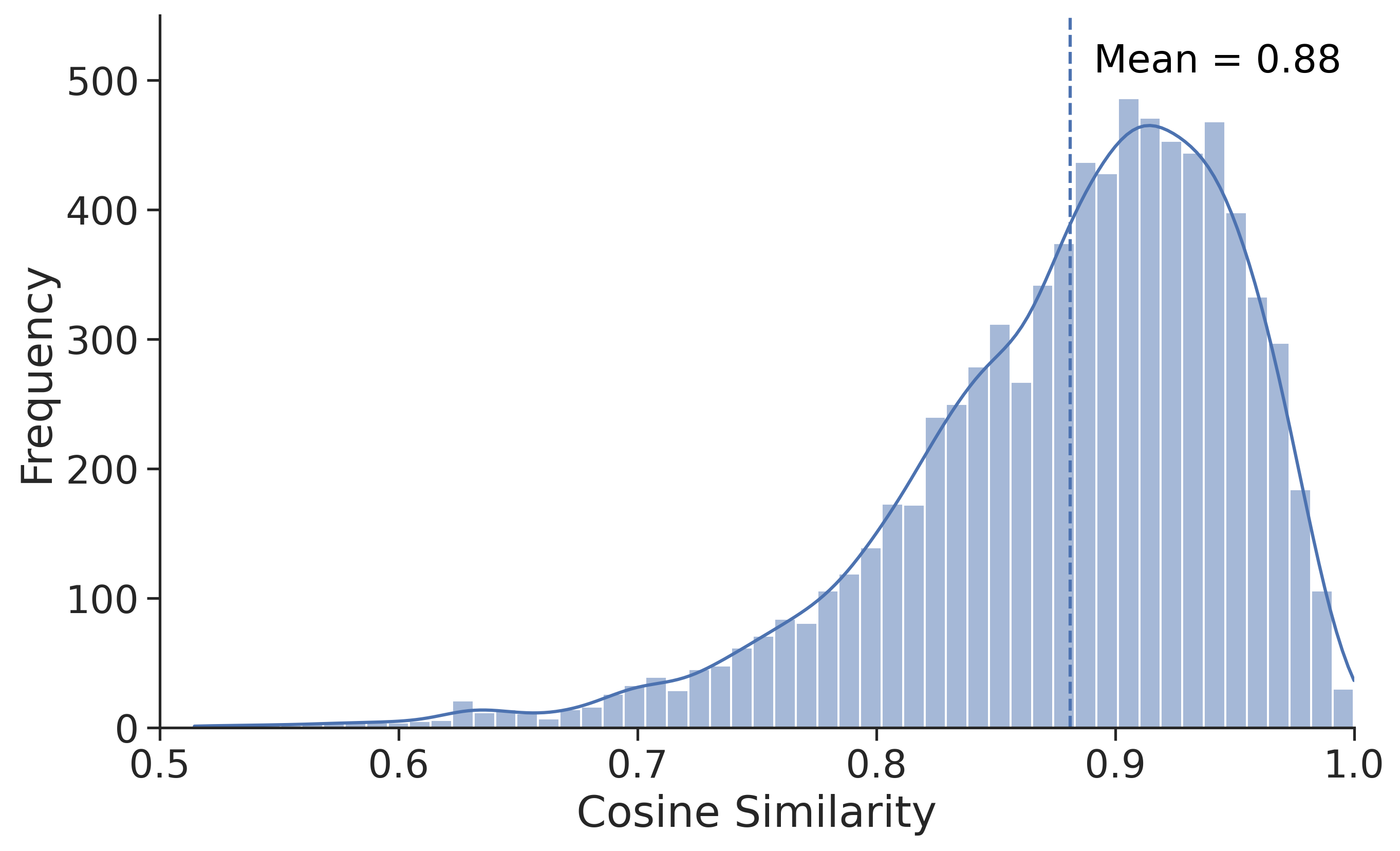}
\caption{Distribution of cosine similarities between Spider questions and generated reformulations.}
\label{fig:cos_sims}

\end{figure}

Our primary training dataset is Spider  \citep{yu-etal-2018-spider}, 
which contains 7,000 questions to 140 different databases and 3,981 target queries.\footnote{We exclude the single-domain datasets \citet{yu-etal-2018-spider} employ in addition to their data.} Although there can be more than one question for the same intent (usually two),  linguistic variations tend to be scanty and limited. We augment Spider with additional questions using ChatGPT as an automatic annotator. For each intent in the original training set, we generate two  question reformulations based on the types specified in Section~\ref{sec:data_generation}. We choose the augmentation types randomly and do not accept duplicates.  
Figure~\ref{fig:cos_sims} shows the distribution of cosine similarities between the original Spider questions and the generated reformulations. 
We measure cosine similarity based on the SimCSE embeddings of \citet{gao-etal-2021-simcse}. As can be seen, the majority of paraphrases are semantically similar to the Spider question (the mean similarity is 0.88). Experiments with different filtering thresholds (ranging from 0.5 to 0.7 with a step of~0.05) revealed that storing all generated examples, effectively adopting a threshold of 0.5, obtained best  results.
Additionally, we manually inspected 100~reformulations and found only~6\%  to be incorrect (i.e.,~inaccurate expressions of intent).  Analysis in Appendix~\ref{app:error_analysis} further shows that our augmentations do not affect the nature of parsing errors.

The resulting training set contains~14,954 instances; statistics for each category are in  Table~\ref{tab:statistics} and examples in Appendix~\ref{sec:examples:augmented}.
The cost of calling the ChatGPT API to obtain our augmentations is approximately~7.5\$.

\subsection{Evaluation Datasets} The Spider development set consists of 1,034 questions to 
 20 databases and 564 target SQL Training Datasets. 
Since  these questions share the same style and level of detail as the training set, we instead focus on evaluation sets with more natural and diverse language. Specifically, we present results on two groups of evaluation sets. The first group are datasets derived from the Spider development set, featuring identical SQL Training Datasets and databases which  allow us to assess the model's resilience to variations in linguistic expression.
The second group are  independent datasets which not only  differ in language usage but also in SQL style and database specifics. This  allows us to  evaluate model performance in more realistic conditions. 

\paragraph{Datasets Based on Spider}
\citet{chang2023drspider} have recently released 
Dr.Spider, a comprehensive robustness benchmark which includes 9~evaluation sets with 7,593 examples of  perturbations in natural language questions (NLQ sets). They have also created 
evaluation sets for  database and SQL perturbations which are out of scope for this work.
NLQ perturbation sets are based on the Spider development set, they  contain the same databases and gold Training Datasets, deviating only in terms of the questions asked.
They are generated with  OPT \citep{zhang2022opt}, a large pretrained language model, and manually filtered by SQL experts.
There are three main categories of perturbations: change one or a few words that refer to SQL keywords (for example, replace the word \textsl{maximum} referring to the \texttt{max} SQL function with \textsl{the largest}), change references to columns (for example, replace
\textsl{name of the countries} referring  to column \texttt{CountryName} with \textsl{which countries}) and change references to database values (for example, replace \textsl{players from the USA} referring to the value \texttt{USA}   with \textsl{American players}). Changes are made by  replacing  words with their synonyms 
or carrier phrases (e.g.,~\textit{name of the countries} and \textit{which countries}). Note that 
our augmentations target solely language variations and do not manipulate gold SQL queries.

\paragraph{Other Datasets}
GeoQuery \citep{zelle1996learning} is a single-domain semantic parsing dataset with questions to a
database of US geography. We use a version with SQL queries as logical forms and query-based splits \citep{finegan-dollak-etal-2018-improving} with a test set of 182 examples. GeoQuery questions are concise and their interpretation often depends on domain knowledge.
For example, in the question
\textsl{what is the largest city in the smallest state in the usa}, \textsl{the largest city} implies the city with the largest population but \textsl{the smallest state} implies the state with the smallest area.

KaggleDBQA \citep{lee-etal-2021-kaggledbqa} is a cross-domain text-to-SQL dataset for testing models under more realistic conditions. It contains 272 examples related to 8 real-world databases which can have abbreviated table and column names and ``dirty'' values. Questions were collected  with annotators having access to column descriptions only, rather than the actual  database schema (the dataset provides these descriptions but we do not use them). This simulates realistic database usage but also creates a challenge for semantic parsers as questions cannot be easily aligned  to target SQL queries. For example, the question
\textit{Which artist/group is most productive?} to a database with information on hip hop torrents should be parsed into query \texttt{SELECT artist FROM torrents GROUP BY artist ORDER BY count(groupName) DESC LIMIT 1}, as \textsl{productive} refers to the number of releases and column \texttt{groupName} contains released titles. 
 
\subsection{Models}
Current approaches frame  text-to-SQL parsing  as a sequence-to-sequence problem. The input is the concatenation of question and database entities, including table and column names, and content values extracted based on  string matching, and the output is an SQL query. \citet{shaw-etal-2021-compositional} show that a pre-trained T5-3B  model \citep{raffel-etal-2020-t5}  fine-tuned on Spider \cite{yu-etal-2018-spider}  is a competitive text-to-SQL parser. \citet{scholak-etal-2021-picard} build on this approach with PICARD, a method for constrained decoding that filters the beam at each generation step, taking into account task-specific constraints such as grammatical correctness and consistency with the database. Recently, \citet{resdsql2023} propose RESDSQL, an approach that decouples schema
linking from SQL parsing. They 
first filter relevant database entities and then use T5-3B  to generate a sketch (i.e.,~SQL keywords) and then the actual SQL query.  We use the best version of their model which also leverages NatSQL intermediate representations  \cite{gan-etal-2021-natural-sql}.

We use the implementations from \citet{scholak-etal-2021-picard} and  \citet{resdsql2023} for training  models on augmented data and their released checkpoints for training on the original Spider. All models are trained for 100~epochs; we use a batch size of~200 for the base T5-3B to reduce   the computational cost, leaving all other hyperparameters unchanged. We train on a single NVIDIA A100 GPU.

Our approach to data augmentation is model agnostic but our experiments focus on settings  where the model is specifically trained 
or fine-tuned on text-to-SQL data. An alternative is large language models which are trained on huge text collections (including code) and able to  translate natural language  to SQL, without further fine-tuning on task-specific data \cite{rajkumar2022evaluating}. Since our augmentations are generated by ChatGPT, a model trained with Reinforcement Learning for Human Feedback \citep{rlhf}, we include it as a standalone baseline. Following \citet{liu2023comprehensive}, we prompt ChatGPT in a \emph{zero-shot} setting  with 
the description of the database schema followed by the question (the full prompt is shown in Appendix~\ref{app:chat_prompt}).
Large language models like ChatGPT differ from task-specific models in many respects, including potential use cases, resource requirements, transparency,  and accessibility and thus any comparison should be interpreted with a grain of salt.

\subsection{Metrics}

We report performance as 
 execution accuracy which compares the execution results of gold and predicted queries (using the implementation of \citealt{zhong-etal-2020-semantic}).

Firstly, we evaluate model \emph{robustness to perturbations} in questions, by considering zero-shot parsing on Dr.Spider. Evaluation sets for Dr.Spider NLQ fall in two categories:
pre-perturbed sets are subsets of the Spider development set, while  post-perturbed sets  are the same subsets but with  rewritten questions instead of the original Spider ones. Execution accuracy on post-perturbed sets measures \textit{absolute} robustness, while the difference in execution accuracy between pre- and \mbox{post-perturbed} sets  measures \textit{relative} robustness.
We also evaluate \emph{out-of-domain generalization} by considering the execution accuracy of zero-shot parsers on GeoQuery and KaggleDBQA.

\section{Results}
Our experiments compare models trained on the original Spider training data against models trained on Spider with our augmentations. We also report results for ChatGPT tested in a 
\mbox{zero-shot} mode. Appendix~\ref{app:add_results} provides additional results (on more evaluation sets) and   
  detailed versions of all tables.

\subsection{Robustness to Question Perturbations}

Table~\ref{tab:robustness} reports average execution accuracy  on evaluation sets from Dr.Spider \citep{chang2023drspider} containing perturbations in natural language questions. We also present results on the original Spider development set. 
Pre/Post refer to subsets before/after perturbations (post-perturbation sets are the same Spider subsets but with the questions rewritten; absolute robustness).

We compare T5-3B  with and without PICARD and RESDSQL models fine-tuned on the original Spider data and our augmentations; we also provide  results for ChatGPT evaluated in the zero-shot setting. 
Our results show that ChatGPT is  most vulnerable to question reformulations among all models.  \citet{chang2023drspider} reach similar conclusions with Codex \citep{chen-codex}, another large pre-trained language model,   and hypothesize  this is due to the training data being biased towards docstrings (which is what most natural language utterances look like on websites like GitHub).

Absolute robustness (accuracy on post-perturbed sets) improves by more than 3\% for augmented models compared to  base models in almost all  cases.  Moreover, the performance gap on pre- and post-perturbed data decreases which  indicates better relative robustness for augmented models.
Augmented RESDSQL delivers the highest post-perturbation accuracy of~72.5\% and augmented PICARD demonstrates the smallest gap between pre- and post-perturbations of 8.4\% confirming that our augmentations improve both absolute and relative robustness.

Augmented models do not have an advantage over base models on 
the original Spider development set (see the last row in Table~\ref{tab:robustness}). There are two reasons for this: firstly, we augment questions only without adding new SQL queries, and secondly, augmentations shift the language distribution by removing specific details and rendering  questions more natural, but the  development set  remains closer to the original training set.

Augmented models do not have an advantage over base models on 
the original Spider development set (see the last row in Table~\ref{tab:robustness}). There are two reasons for this: firstly, we augment questions only without adding new SQL queries, and secondly, augmentations shift the language distribution by removing specific details and rendering  questions more natural, but the  development set  remains closer to the original training set.

\begin{table}[t]
\centering
\small{
\begin{tabular}{@{}l@{\;\;\;}cc@{\;\;\;}c@{\;\;\;}r}
\toprule
& \multirow{2}{*}{} & \multicolumn{3}{c}{Dr.Spider NLQ} \\  \cmidrule(lr){3-5}
\multicolumn{1}{c}{Model} & Spider Dev $\uparrow$ & Pre $\uparrow$ & Post $\uparrow$ & Diff $\downarrow$ \\\midrule

T5-3B &  74.4 & 70.3 & 58.9 & 11.4 \\\addlinespace[3pt]
\quad +Augmented & 75.3 & 72.6 & 62.7 & 9.9  \\\addlinespace[6pt]
PICARD & 79.3 & 76.0 & 65.0 & 11.0 \\\addlinespace[3pt]
\quad +Augmented & 79.3 & 76.7 & 68.3 & \textbf{8.4}  \\\addlinespace[6pt]

RESDSQL & \textbf{84.1} & \textbf{84.7} & 69.3 & 15.4  \\\addlinespace[3pt]
\quad +Augmented & 84.0 & 84.6 & \textbf{72.5} & 12.1  \\\midrule
ChatGPT & 72.2 & 73.8 & 57.9 &	15.9 
\\ \bottomrule
\end{tabular}
}
\caption{Execution Accuracy on Spider development set and Dr.Spider NLQ subsets, before  (Pre) and after perturbations (Post: absolute robustness) and the gap between them (Diff: relative robustness). All models are tested in a zero-shot setting;  +Augmented refers to  models fine-tuned on the augmented Spider data.}
\label{tab:robustness}
\end{table}

\begin{table*}[h!]
\centering
\footnotesize
\begin{tabular}{lcc@{\;\;}c@{\;\;}c@{\;\;}c@{\;\;}c@{\;\;}c@{\;\;}c@{\;\;}c@{\;\;\;}c}
\toprule
 &  & \multicolumn{9}{c}{KaggleDBQA} \\\addlinespace[3pt]
\cmidrule(lr){3-11}
 \multicolumn{1}{c}{Model} &  GeoQuery & Nuclear & Crime & Pesticide & Math
& Baseball & Fires & WhatCD & Soccer  & Avg \\\midrule
T5-3B &  54.4  & 59.4 & 48.2 & 16.0 & 7.1 & 20.5 & 43.2 & 7.3 & 16.7 & 27.3
\\\addlinespace[3pt]
\quad +Augmented &  60.4  & 56.3 & 48.2 & 18.0 & 7.1 & 20.5 & \textbf{43.2} & \textbf{26.8} & 22.2 & 30.3
\\\addlinespace[3pt]
PICARD &  56.6  & 59.4 & \textbf{51.9} & 18.0 & 10.7 & \textbf{25.6} & \textbf{43.2} & 9.8 & 22.2 & 30.1
\\\addlinespace[3pt]
\quad +Augmented &  \textbf{62.6} & 56.3 & 48.1 & 22.0 & 14.3 & \textbf{25.6} & \textbf{43.2} & 24.4 & 27.8 & 32.7
\\\addlinespace[3pt]
RESDSQL &  56.6 & 59.4 & 48.1 & 16.0 & \textbf{25.0} & 23.1 & \textbf{43.2} & 17.1 & 22.2 & 31.8
\\\addlinespace[3pt]
\quad +Augmented & 59.3 & \textbf{65.6} & 44.4 & \textbf{24.0} & \textbf{25.0} & 23.1 & \textbf{43.2} & 19.5 & \textbf{27.8} & \textbf{34.1}
\\\addlinespace[3pt]
ChatGPT & 20.9 & 34.4 & 18.5 & 16.0 & 10.7 & 15.4 & 27.0 & 4.9 & 16.7 & 17.9
\\\bottomrule
\end{tabular}
\caption{Execution accuracy on GeoQuery  test set  (query splits) and  different databases from KaggleDBQA. All models are tested in a zero-shot setting;  +Augmented refers to  models fine-tuned on the augmented Spider data.}
\label{tab:out-of-domain}
\end{table*}

\subsection{Generalization to Other Datasets}

Table~\ref{tab:out-of-domain} summarizes our results in the 
more challenging zero-shot setting. Specifically, we evaluate
model performance on two out-of-domain datasets, namely
GeoQuery \cite{zelle1996learning} and KaggleDBQA \cite{lee-etal-2021-kaggledbqa}.  Both datasets differ from Spider in many respects, i.e., the types of  questions being asked, the style of SQL queries, and the database structure. 

We find ChatGPT performs very poorly on these datasets compared to  models fine-tuned on Spider with or without augmentations.
In all cases, augmented models improve execution accuracy compared to base models. PICARD trained with augmentations performs best on GeoQuery reaching an accuracy of~62.6\% (a~6\% difference against the base model).
Augmented RESDSQL performs best on KaggleDBQA, which is more challenging, 
reaching an average accuracy of~34.1\%.
Augmentations are generally helpful but not across all individual categories (note that categories are represented by a limited number of examples per database and even  a small number of errors can result in a drop of several percentage points). We suspect the low accuracy on KaggleDBQA is primarily due to  challenges that are  unrelated to language variation. In particular, its databases contain abbreviations which might be difficult to parse and SQL queries exemplify operations which are not present in Spider (e.g.,~arithmetic operators between columns).

\subsection{Ablations and Analysis}
We next investigate the impact of different types of question reformulations introduced in  Section~\ref{sec:data_generation}, and also compare against related augmentation methods:  \citet{gan-etal-2021-towards} manually annotate Spider-Syn with synonym substitutions, whereas  \citet{ma_and_wang2021} introduce \textsc{MT-teql}, a framework for  generating semantics-preserving variants of  
utterances and database schemas. We  use a version of \textsc{MT-teql}  that changes prefixes and aggregator mentions in  Spider questions. Additionally, we include a baseline which follows our procedure for data generation but uses only one prompt: provide \textit{different ways of expressing} a question.

Table~\ref{tab:ablation} shows the execution accuracy of  \mbox{T5-3B}  trained with  and without augmentations  pertaining to Deletion, Substitution, Rewriting, and Paraphrasing. We also include results with All augmentations combined. 
The ablation study shows that  different types of augmentation  are helpful for different  datasets. On GeoQuery,  models augmented with deletions and substitutions perform best; substitutions  also perform best on the NLQ sets of Dr.Spider and KaggleDBQA. Paraphrasing-based augmentations are best for   the original Spider development set,  with Rewriting  trailing behind. 
Results obtained with a single prompt (\texttt{express in a different way}) further  
illustrate the need for diverse instructions.
 We also trained T5-3B  with augmentations from Spider-Syn \citep{gan-etal-2021-towards} and  \textsc{MT-teql} \citep{ma_and_wang2021}. For a fair comparison, we randomly sample \textsc{MT-teql} examples with question transformations to 
match the training size obtained through our augmentations (Spider-Syn and one-prompt baselines also match our training size). As can be seen
in Table~\ref{tab:ablation}, our combined augmentations outperform models trained on Spider-Syn  and  \textsc{MT-teql} on all  evaluation sets (Dr.Spider NLG, GeoQuery, and KaggleDBQA) and the improvement comes from  reformulating the questions rather than increasing the training set.  
 
\begin{table*}[t]
\centering
\footnotesize
\begin{tabular}{@{}lccc@{\;\;\;}ccc}
\toprule
& \multirow{2}{*}{} & \multicolumn{2}{c}{Dr.Spider NLQ} & \multirow{2}{*}{} & \multirow{2}{*}{} \\\cmidrule(lr){3-4}
\multicolumn{1}{c}{Model} & Spider Dev $\uparrow$ & Post $\uparrow$  & Diff $\downarrow$  &  GeoQuery $\uparrow$  &   KaggleDBQA $\uparrow$ \\\midrule
T5-3B  & 74.4 & 58.9 & 11.4 & 54.4 & 27.3 \\\addlinespace[3pt]
\quad + Deletion & 74.7 & 59.7 & 11.5  & 56.0 &  28.7 
\\\addlinespace[3pt]
\quad + Substitution & 75.1 & 62.9 & \textbf{9.8} &  56.0 & \textbf{31.2} 
\\\addlinespace[3pt]
\quad + Rewriting  & 75.0 & 62.3 & 11.2 & 53.8 & 27.4 
\\\addlinespace[3pt]
\quad + Paraphrase  & 75.3 & 61.4 & 11.6 & 41.8 & 25.9 
\\\addlinespace[3pt]
\quad + All (ours)  & 75.3 & \textbf{63.2} & 9.9 & \textbf{60.4}  & 30.3
\\\midrule
\quad + One Prompt   & 74.4 & {60.4} & 15.4 & {40.7}  & {29.2}  \\\addlinespace[3pt]
\quad + Spider-Syn  & \textbf{75.6} & 59.2 & 13.2 & 49.5& 27.0\\\addlinespace[3pt]
\quad + \textsc{MT-teql}* & 75.0 & 62.0 & 11.1 & 47.8 & 29.2
\\\bottomrule
\end{tabular}
\caption{Execution accuracy on  Spider development set,  Dr.Spider NLQ (Post: absolute robustness; Diff:  relative  robustness), GeoQuery, and KaggleDBQA  for T5-3B base and trained with different augmentations including Spider-Syn \citep{gan-etal-2021-towards} and sub-sampled (diacritic~*) version of \textsc{MT-teql} \citep{ma_and_wang2021}.
}
\label{tab:ablation}
\end{table*}
The results in Table~\ref{tab:ablation}  reaffirm the observation
that different evaluation sets exemplify different linguistic variations
and that there is no single type of augmentation that represents them all. Rather, a \emph{combination} of augmentations is needed to perform well
\emph{across} datasets. This in turn  suggests that a  model can acquire
 useful knowledge by being exposed to  a \emph{diverse} range of linguistic variations. We also observe that a model trained on combined augmentations outperforms models trained on more specialized datasets (i.e., Spider-Syn and \textsc{MT-teql})  which confirms that relying solely on local transformations of the questions is not sufficient for better generalization. 

\section{Conclusion}
We propose to enhance the generalization capabilities of text-to-SQL parsers by increasing natural language variation in the training data.  We leverage a large language model like ChatGPT to automatically generate a variety of question reformulations, thereby augmenting existing datasets with more natural and diverse questions. We evaluate state-of-the-art models trained with and without our augmentations on a variety of challenging datasets focusing on robustness (to perturbations) and out-of-domain generalization. Across models and datasets we find that augmentations improve performance by a wide margin. Our experiments further underscore the need for a broad range of augmentations representing the full spectrum of rewrite operations. In the future, we plan to explore the potential of large language   models for multilingual semantic parsing.

\section*{Limitations}  
Our work aims to increase the robustness of semantic parsers against  natural language variation but does not handle problems related to SQL queries and database structures that are also important for out-of-domain generalization. We obtain augmentations using ChatGPT, a black-box model provided by OpenAI, which limits its usage for non-academic purposes.
Our augmentations are unfiltered and may add a small amount of noise to training data. Moreover,  even though our proposed rewrite operations are diverse,  they may still not cover all possible reformulations. In fact, we found it challenging for ChatGPT to generate wildly different expressions of the original intent. Finally, this work does not consider multilingual or conversational semantic parsing which we hope to explore in the future.

\subsection*{Acknowledgments}
We thank the meta-reviewer and anonymous reviewers for their constructive feedback. The authors also thank Hao Zheng for insightful comments on earlier versions of this work. We gratefully acknowledge the support of
the UK Engineering and Physical Sciences Research Council (grant EP/L016427/1).

\bibliography{anthology,custom}

\begin{thebibliography}{41}
\expandafter\ifx\csname natexlab\endcsname\relax\def\natexlab#1{#1}\fi

\bibitem[{Artzi and Zettlemoyer(2011)}]{artzi-zettlemoyer-2011-bootstrapping}
Yoav Artzi and Luke Zettlemoyer. 2011.
\newblock \href {https://aclanthology.org/D11-1039} {Bootstrapping semantic parsers from conversations}.
\newblock In \emph{Proceedings of the 2011 Conference on Empirical Methods in Natural Language Processing}, pages 421--432, Edinburgh, Scotland, UK. Association for Computational Linguistics.

\bibitem[{Brown et~al.(2020)Brown, Mann, Ryder, Subbiah, Kaplan, Dhariwal, Neelakantan, Shyam, Sastry, Askell, Agarwal, Herbert-Voss, Krueger, Henighan, Child, Ramesh, Ziegler, Wu, Winter, Hesse, Chen, Sigler, Litwin, Gray, Chess, Clark, Berner, McCandlish, Radford, Sutskever, and Amodei}]{brown2020language}
Tom Brown, Benjamin Mann, Nick Ryder, Melanie Subbiah, Jared~D Kaplan, Prafulla Dhariwal, Arvind Neelakantan, Pranav Shyam, Girish Sastry, Amanda Askell, Sandhini Agarwal, Ariel Herbert-Voss, Gretchen Krueger, Tom Henighan, Rewon Child, Aditya Ramesh, Daniel Ziegler, Jeffrey Wu, Clemens Winter, Chris Hesse, Mark Chen, Eric Sigler, Mateusz Litwin, Scott Gray, Benjamin Chess, Jack Clark, Christopher Berner, Sam McCandlish, Alec Radford, Ilya Sutskever, and Dario Amodei. 2020.
\newblock \href {https://proceedings.neurips.cc/paper_files/paper/2020/file/1457c0d6bfcb4967418bfb8ac142f64a-Paper.pdf} {Language models are few-shot learners}.
\newblock In \emph{Proceedings of the 33st Annual Conference on Neural Information Processing Systems}, volume~33, pages 1877--1901. Curran Associates, Inc.

\bibitem[{Campagna et~al.(2017)Campagna, Ramesh, Xu, Fischer, and Lam}]{10.1145/3038912.3052562}
Giovanni Campagna, Rakesh Ramesh, Silei Xu, Michael Fischer, and Monica~S. Lam. 2017.
\newblock \href {https://doi.org/10.1145/3038912.3052562} {Almond: The architecture of an open, crowdsourced, privacy-preserving, programmable virtual assistant}.
\newblock In \emph{Proceedings of the 26th International Conference on World Wide Web}, WWW '17, page 341–350, Republic and Canton of Geneva, CHE. International World Wide Web Conferences Steering Committee.

\bibitem[{Chang et~al.(2023)Chang, Wang, Dong, Pan, Zhu, Li, Lan, Zhang, Jiang, Lilien, Ash, Wang, Wang, Castelli, Ng, and Xiang}]{chang2023drspider}
Shuaichen Chang, Jun Wang, Mingwen Dong, Lin Pan, Henghui Zhu, Alexander~Hanbo Li, Wuwei Lan, Sheng Zhang, Jiarong Jiang, Joseph Lilien, Steve Ash, William~Yang Wang, Zhiguo Wang, Vittorio Castelli, Patrick Ng, and Bing Xiang. 2023.
\newblock \href {https://openreview.net/forum?id=Wc5bmZZU9cy} {Dr.spider: A diagnostic evaluation benchmark towards text-to-{SQL} robustness}.
\newblock In \emph{The 11th International Conference on Learning Representations}.

\bibitem[{Chen et~al.(2021)Chen, Tworek, Jun, Yuan, Pinto, Kaplan, Edwards, Burda, Joseph, Brockman, Ray, Puri, Krueger, Petrov, Khlaaf, Sastry, Mishkin, Chan, Gray, Ryder, Pavlov, Power, Kaiser, Bavarian, Winter, Tillet, Such, Cummings, Plappert, Chantzis, Barnes, Herbert-Voss, Guss et~al.}]{chen-codex}
Mark Chen, Jerry Tworek, Heewoo Jun, Qiming Yuan, Henrique Ponde de~Oliveira Pinto, Jared Kaplan, Harri Edwards, Yuri Burda, Nicholas Joseph, Greg Brockman, Alex Ray, Raul Puri, Gretchen Krueger, Michael Petrov, Heidy Khlaaf, Girish Sastry, Pamela Mishkin, Brooke Chan, Scott Gray, Nick Ryder, Mikhail Pavlov, Alethea Power, Lukasz Kaiser, Mohammad Bavarian, Clemens Winter, Philippe Tillet, Felipe~Petroski Such, Dave Cummings, Matthias Plappert, Fotios Chantzis, Elizabeth Barnes, Ariel Herbert-Voss, Guss, et~al. 2021.
\newblock \href {http://arxiv.org/abs/2107.03374} {Evaluating large language models trained on code}.

\bibitem[{Chowdhery et~al.(2022)Chowdhery, Narang, Devlin, Bosma, Mishra, Roberts, Barham, Chung, Sutton, Gehrmann et~al.}]{chowdhery2022palm}
Aakanksha Chowdhery, Sharan Narang, Jacob Devlin, Maarten Bosma, Gaurav Mishra, Adam Roberts, Paul Barham, Hyung~Won Chung, Charles Sutton, Sebastian Gehrmann, et~al. 2022.
\newblock \href {http://arxiv.org/abs/2204.02311} {Palm: Scaling language modeling with pathways}.

\bibitem[{Christiano et~al.(2017)Christiano, Leike, Brown, Martic, Legg, and Amodei}]{rlhf}
Paul~F Christiano, Jan Leike, Tom Brown, Miljan Martic, Shane Legg, and Dario Amodei. 2017.
\newblock \href {https://proceedings.neurips.cc/paper_files/paper/2017/file/d5e2c0adad503c91f91df240d0cd4e49-Paper.pdf} {Deep reinforcement learning from human preferences}.
\newblock In \emph{Proceedings of the 31st Annual Conference on Neural Information Processing Systems}, volume~30, Long Beach, CA, USA. Curran Associates, Inc.

\bibitem[{Dai et~al.(2023)Dai, Liu, Liao, Huang, Cao, Wu, Zhao, Xu, Liu, Liu, Li, Zhu, Cai, Sun, Li, Shen, Liu, and Li}]{dai2023auggpt}
Haixing Dai, Zhengliang Liu, Wenxiong Liao, Xiaoke Huang, Yihan Cao, Zihao Wu, Lin Zhao, Shaochen Xu, Wei Liu, Ninghao Liu, Sheng Li, Dajiang Zhu, Hongmin Cai, Lichao Sun, Quanzheng Li, Dinggang Shen, Tianming Liu, and Xiang Li. 2023.
\newblock \href {http://arxiv.org/abs/2302.13007} {Auggpt: Leveraging chatgpt for text data augmentation}.

\bibitem[{Deng et~al.(2021)Deng, Awadallah, Meek, Polozov, Sun, and Richardson}]{deng-etal-2021-structure}
Xiang Deng, Ahmed~Hassan Awadallah, Christopher Meek, Oleksandr Polozov, Huan Sun, and Matthew Richardson. 2021.
\newblock \href {https://doi.org/10.18653/v1/2021.naacl-main.105} {Structure-grounded pretraining for text-to-{SQL}}.
\newblock In \emph{Proceedings of the 2021 Conference of the North American Chapter of the Association for Computational Linguistics: Human Language Technologies}, pages 1337--1350, Online. Association for Computational Linguistics.

\bibitem[{Dukes(2014)}]{dukes-2014-semeval}
Kais Dukes. 2014.
\newblock \href {https://doi.org/10.3115/v1/S14-2006} {{S}em{E}val-2014 task 6: Supervised semantic parsing of robotic spatial commands}.
\newblock In \emph{Proceedings of the 8th International Workshop on Semantic Evaluation ({S}em{E}val 2014)}, pages 45--53, Dublin, Ireland. Association for Computational Linguistics.

\bibitem[{Finegan-Dollak et~al.(2018)Finegan-Dollak, Kummerfeld, Zhang, Ramanathan, Sadasivam, Zhang, and Radev}]{finegan-dollak-etal-2018-improving}
Catherine Finegan-Dollak, Jonathan~K. Kummerfeld, Li~Zhang, Karthik Ramanathan, Sesh Sadasivam, Rui Zhang, and Dragomir Radev. 2018.
\newblock \href {https://doi.org/10.18653/v1/P18-1033} {Improving text-to-{SQL} evaluation methodology}.
\newblock In \emph{Proceedings of the 56th Annual Meeting of the Association for Computational Linguistics (Volume 1: Long Papers)}, pages 351--360, Melbourne, Australia. Association for Computational Linguistics.

\bibitem[{Gan et~al.(2021{\natexlab{a}})Gan, Chen, Huang, Purver, Woodward, Xie, and Huang}]{gan-etal-2021-towards}
Yujian Gan, Xinyun Chen, Qiuping Huang, Matthew Purver, John~R. Woodward, Jinxia Xie, and Pengsheng Huang. 2021{\natexlab{a}}.
\newblock \href {https://doi.org/10.18653/v1/2021.acl-long.195} {Towards robustness of text-to-{SQL} models against synonym substitution}.
\newblock In \emph{Proceedings of the 59th Annual Meeting of the Association for Computational Linguistics and the 11th International Joint Conference on Natural Language Processing (Volume 1: Long Papers)}, pages 2505--2515, Online. Association for Computational Linguistics.

\bibitem[{Gan et~al.(2021{\natexlab{b}})Gan, Chen, and Purver}]{gan-etal-2021-exploring}
Yujian Gan, Xinyun Chen, and Matthew Purver. 2021{\natexlab{b}}.
\newblock \href {https://doi.org/10.18653/v1/2021.emnlp-main.702} {Exploring underexplored limitations of cross-domain text-to-{SQL} generalization}.
\newblock In \emph{Proceedings of the 2021 Conference on Empirical Methods in Natural Language Processing}, pages 8926--8931, Online and Punta Cana, Dominican Republic. Association for Computational Linguistics.

\bibitem[{Gan et~al.(2021{\natexlab{c}})Gan, Chen, Xie, Purver, Woodward, Drake, and Zhang}]{gan-etal-2021-natural-sql}
Yujian Gan, Xinyun Chen, Jinxia Xie, Matthew Purver, John~R. Woodward, John Drake, and Qiaofu Zhang. 2021{\natexlab{c}}.
\newblock \href {https://doi.org/10.18653/v1/2021.findings-emnlp.174} {Natural {SQL}: Making {SQL} easier to infer from natural language specifications}.
\newblock In \emph{Findings of the Association for Computational Linguistics: EMNLP 2021}, pages 2030--2042, Punta Cana, Dominican Republic. Association for Computational Linguistics.

\bibitem[{Gao et~al.(2021)Gao, Yao, and Chen}]{gao-etal-2021-simcse}
Tianyu Gao, Xingcheng Yao, and Danqi Chen. 2021.
\newblock \href {https://doi.org/10.18653/v1/2021.emnlp-main.552} {{S}im{CSE}: Simple contrastive learning of sentence embeddings}.
\newblock In \emph{Proceedings of the 2021 Conference on Empirical Methods in Natural Language Processing}, pages 6894--6910, Online and Punta Cana, Dominican Republic. Association for Computational Linguistics.

\bibitem[{Hazoom et~al.(2021)Hazoom, Malik, and Bogin}]{hazoom-etal-2021-text}
Moshe Hazoom, Vibhor Malik, and Ben Bogin. 2021.
\newblock \href {https://doi.org/10.18653/v1/2021.nlp4prog-1.9} {Text-to-{SQL} in the wild: A naturally-occurring dataset based on stack exchange data}.
\newblock In \emph{Proceedings of the 1st Workshop on Natural Language Processing for Programming (NLP4Prog 2021)}, pages 77--87, Online. Association for Computational Linguistics.

\bibitem[{He et~al.(2023)He, Lin, Gong, Jin, Zhang, Lin, Jiao, Yiu, Duan, and Chen}]{he2023annollm}
Xingwei He, Zhenghao Lin, Yeyun Gong, A-Long Jin, Hang Zhang, Chen Lin, Jian Jiao, Siu~Ming Yiu, Nan Duan, and Weizhu Chen. 2023.
\newblock \href {http://arxiv.org/abs/2303.16854} {Annollm: Making large language models to be better crowdsourced annotators}.

\bibitem[{Huang et~al.(2021)Huang, Li, Qu, and Pan}]{Huang-etal-2021-robustness}
Shuo Huang, Zhuang Li, Lizhen Qu, and Lei Pan. 2021.
\newblock \href {https://doi.org/10.18653/v1/2021.eacl-main.292} {On robustness of neural semantic parsers}.
\newblock In \emph{Proceedings of the 16th Conference of the European Chapter of the Association for Computational Linguistics: Main Volume}, pages 3333--3342, Online. Association for Computational Linguistics.

\bibitem[{Jiang et~al.(2022)Jiang, Hu, Lan, Zhu, Chauhan, Li, Pan, Wang, Hang, Zhang, Dong, Lilien, Ng, Wang, Castelli, and Xiang}]{Jiang2022}
Jiarong Jiang, Yiqun Hu, Wuwei Lan, Henry Zhu, Anuj Chauhan, Alexander Li, Lin Pan, Jun Wang, Chung-Wei Hang, Sheng Zhang, Marvin Dong, Joe Lilien, Patrick Ng, Zhiguo Wang, Vittorio Castelli, and Bing Xiang. 2022.
\newblock \href {https://openreview.net/pdf?id=9Ue2wCnvdMy} {Importance of synthesizing high-quality data for text-to-sql parsing}.
\newblock In \emph{NeurIPS 2022 Workshop on SyntheticData4ML}.

\bibitem[{Lee et~al.(2021)Lee, Polozov, and Richardson}]{lee-etal-2021-kaggledbqa}
Chia-Hsuan Lee, Oleksandr Polozov, and Matthew Richardson. 2021.
\newblock \href {https://doi.org/10.18653/v1/2021.acl-long.176} {{K}aggle{DBQA}: Realistic evaluation of text-to-{SQL} parsers}.
\newblock In \emph{Proceedings of the 59th Annual Meeting of the Association for Computational Linguistics and the 11th International Joint Conference on Natural Language Processing (Volume 1: Long Papers)}, pages 2261--2273, Online. Association for Computational Linguistics.

\bibitem[{Li et~al.(2023{\natexlab{a}})Li, Zhang, Li, and Chen}]{resdsql2023}
Haoyang Li, Jing Zhang, Cuiping Li, and Hong Chen. 2023{\natexlab{a}}.
\newblock \href {https://arxiv.org/pdf/2302.05965.pdf} {Resdsql: Decoupling schema linking and skeleton parsing for text-to-sql}.
\newblock In \emph{Proceedings of the 37th AAAI Conference on Artificial Intelligence}, pages 13067--13075, Washington, DC, USA. AAAI Press.

\bibitem[{Li et~al.(2019)Li, Wang, Ku, Tian, and Wang}]{Li:ea:2019}
Jingjing Li, Wenlu Wang, Wei-Shinn Ku, Yingtao Tian, and Haixun Wang. 2019.
\newblock \href {https://doi.org/10.1145/3347146.3359069} {Spatialnli: A spatial domain natural language interface to databases using spatial comprehension}.
\newblock In \emph{Proceedings of the 27th ACM SIGSPATIAL International Conference on Advances in Geographic Information Systems}, SIGSPATIAL '19, page 339–348, New York, NY, USA. Association for Computing Machinery.

\bibitem[{Li et~al.(2023{\natexlab{b}})Li, Hui, Qu, Li, Yang, Li, Wang, Qin, Cao, Geng, Huo, Ma, Chang, Huang, Cheng, and Li}]{li2023llm}
Jinyang Li, Binyuan Hui, Ge~Qu, Binhua Li, Jiaxi Yang, Bowen Li, Bailin Wang, Bowen Qin, Rongyu Cao, Ruiying Geng, Nan Huo, Chenhao Ma, Kevin C.~C. Chang, Fei Huang, Reynold Cheng, and Yongbin Li. 2023{\natexlab{b}}.
\newblock \href {http://arxiv.org/abs/2305.03111} {Can llm already serve as a database interface? a big bench for large-scale database grounded text-to-sqls}.

\bibitem[{Liu et~al.(2023)Liu, Hu, Wen, and Yu}]{liu2023comprehensive}
Aiwei Liu, Xuming Hu, Lijie Wen, and Philip~S. Yu. 2023.
\newblock \href {http://arxiv.org/abs/2303.13547} {A comprehensive evaluation of chatgpt’s zero-shot text-to-sql capability}.

\bibitem[{Ma and Wang(2021)}]{ma_and_wang2021}
Pingchuan Ma and Shuai Wang. 2021.
\newblock \href {https://doi.org/10.14778/3494124.3494139} {Mt-teql: Evaluating and augmenting neural nlidb on real-world linguistic and schema variations}.
\newblock \emph{Proceedings of the VLDB Endowment}, 15(3):569–582.

\bibitem[{{\H{O}}zcan et~al.(2020){\H{O}}zcan, Quamar, Sen, Lei, and Efthymiou}]{Ozcan:ea:2020}
Fatma {\H{O}}zcan, Abdul Quamar, Jaydeep Sen, Chuan Lei, and Vasilis Efthymiou. 2020.
\newblock \href {https://doi.org/10.1145/3318464.3383128} {State of the art and open challenges in natural language interfaces to data}.
\newblock In \emph{Proceedings of the 2020 ACM SIGMOD International Conference on Management of Data}, SIGMOD '20, page 2629–2636, New York, NY, USA. Association for Computing Machinery.

\bibitem[{Pi et~al.(2022)Pi, Wang, Gao, Guo, Li, and Lou}]{pi-etal-2022-towards}
Xinyu Pi, Bing Wang, Yan Gao, Jiaqi Guo, Zhoujun Li, and Jian-Guang Lou. 2022.
\newblock \href {https://doi.org/10.18653/v1/2022.acl-long.142} {Towards robustness of text-to-{SQL} models against natural and realistic adversarial table perturbation}.
\newblock In \emph{Proceedings of the 60th Annual Meeting of the Association for Computational Linguistics (Volume 1: Long Papers)}, pages 2007--2022, Dublin, Ireland. Association for Computational Linguistics.

\bibitem[{Radhakrishnan et~al.(2020)Radhakrishnan, Srikantan, and Lin}]{Radhakrishnan-etal-2020-COLLOQL}
Karthik Radhakrishnan, Arvind Srikantan, and Xi~Victoria Lin. 2020.
\newblock \href {https://doi.org/10.18653/v1/2020.intexsempar-1.5} {{C}ollo{QL}: Robust text-to-{SQL} over search queries}.
\newblock In \emph{Proceedings of the First Workshop on Interactive and Executable Semantic Parsing}, pages 34--45, Online. Association for Computational Linguistics.

\bibitem[{Raffel et~al.(2020)Raffel, Shazeer, Roberts, Lee, Narang, Matena, Zhou, Li, and Liu}]{raffel-etal-2020-t5}
Colin Raffel, Noam Shazeer, Adam Roberts, Katherine Lee, Sharan Narang, Michael Matena, Yanqi Zhou, Wei Li, and Peter~J. Liu. 2020.
\newblock \href {http://jmlr.org/papers/v21/20-074.html} {Exploring the limits of transfer learning with a unified text-to-text transformer}.
\newblock \emph{Journal of Machine Learning Research}, 21(140):1--67.

\bibitem[{Rajkumar et~al.(2022)Rajkumar, Li, and Bahdanau}]{rajkumar2022evaluating}
Nitarshan Rajkumar, Raymond Li, and Dzmitry Bahdanau. 2022.
\newblock \href {http://arxiv.org/abs/2204.00498} {Evaluating the text-to-sql capabilities of large language models}.

\bibitem[{Scholak et~al.(2021)Scholak, Schucher, and Bahdanau}]{scholak-etal-2021-picard}
Torsten Scholak, Nathan Schucher, and Dzmitry Bahdanau. 2021.
\newblock \href {https://doi.org/10.18653/v1/2021.emnlp-main.779} {{PICARD}: Parsing incrementally for constrained auto-regressive decoding from language models}.
\newblock In \emph{Proceedings of the 2021 Conference on Empirical Methods in Natural Language Processing}, pages 9895--9901, Online and Punta Cana, Dominican Republic. Association for Computational Linguistics.

\bibitem[{Shaw et~al.(2021)Shaw, Chang, Pasupat, and Toutanova}]{shaw-etal-2021-compositional}
Peter Shaw, Ming-Wei Chang, Panupong Pasupat, and Kristina Toutanova. 2021.
\newblock \href {https://doi.org/10.18653/v1/2021.acl-long.75} {Compositional generalization and natural language variation: Can a semantic parsing approach handle both?}
\newblock In \emph{Proceedings of the 59th Annual Meeting of the Association for Computational Linguistics and the 11th International Joint Conference on Natural Language Processing (Volume 1: Long Papers)}, pages 922--938, Online. Association for Computational Linguistics.

\bibitem[{Suhr et~al.(2020)Suhr, Chang, Shaw, and Lee}]{suhr-etal-2020-exploring}
Alane Suhr, Ming-Wei Chang, Peter Shaw, and Kenton Lee. 2020.
\newblock \href {https://doi.org/10.18653/v1/2020.acl-main.742} {Exploring unexplored generalization challenges for cross-database semantic parsing}.
\newblock In \emph{Proceedings of the 58th Annual Meeting of the Association for Computational Linguistics}, pages 8372--8388, Online. Association for Computational Linguistics.

\bibitem[{Wang et~al.(2021)Wang, Yin, Lin, and Xiong}]{wang-etal-2021-learning-synthesize}
Bailin Wang, Wenpeng Yin, Xi~Victoria Lin, and Caiming Xiong. 2021.
\newblock \href {https://doi.org/10.18653/v1/2021.naacl-main.220} {Learning to synthesize data for semantic parsing}.
\newblock In \emph{Proceedings of the 2021 Conference of the North American Chapter of the Association for Computational Linguistics: Human Language Technologies}, pages 2760--2766, Online. Association for Computational Linguistics.

\bibitem[{Wu et~al.(2021)Wu, Wang, Li, Zhang, Xiao, Wu, Zhang, and Wang}]{wu-etal-2021-data}
Kun Wu, Lijie Wang, Zhenghua Li, Ao~Zhang, Xinyan Xiao, Hua Wu, Min Zhang, and Haifeng Wang. 2021.
\newblock \href {https://doi.org/10.18653/v1/2021.emnlp-main.707} {Data augmentation with hierarchical {SQL}-to-question generation for cross-domain text-to-{SQL} parsing}.
\newblock In \emph{Proceedings of the 2021 Conference on Empirical Methods in Natural Language Processing}, pages 8974--8983, Online and Punta Cana, Dominican Republic. Association for Computational Linguistics.

\bibitem[{Yu et~al.(2018)Yu, Zhang, Yang, Yasunaga, Wang, Li, Ma, Li, Yao, Roman, Zhang, and Radev}]{yu-etal-2018-spider}
Tao Yu, Rui Zhang, Kai Yang, Michihiro Yasunaga, Dongxu Wang, Zifan Li, James Ma, Irene Li, Qingning Yao, Shanelle Roman, Zilin Zhang, and Dragomir Radev. 2018.
\newblock \href {https://doi.org/10.18653/v1/D18-1425} {{S}pider: A large-scale human-labeled dataset for complex and cross-domain semantic parsing and text-to-{SQL} task}.
\newblock In \emph{Proceedings of the 2018 Conference on Empirical Methods in Natural Language Processing}, pages 3911--3921, Brussels, Belgium. Association for Computational Linguistics.

\bibitem[{Zelle and Mooney(1996)}]{zelle1996learning}
John~M Zelle and Raymond~J Mooney. 1996.
\newblock Learning to parse database queries using inductive logic programming.
\newblock In \emph{Proceedings of the 13th AAAI Conference on Artificial Intelligence}, volume~2, pages 1050--1055, Portland, Oregon. AAAI Press.

\bibitem[{Zhang et~al.(2022)Zhang, Roller, Goyal, Artetxe, Chen, Chen, Dewan, Diab, Li, Lin, Mihaylov, Ott, Shleifer, Shuster, Simig, Koura, Sridhar, Wang, and Zettlemoyer}]{zhang2022opt}
Susan Zhang, Stephen Roller, Naman Goyal, Mikel Artetxe, Moya Chen, Shuohui Chen, Christopher Dewan, Mona Diab, Xian Li, Xi~Victoria Lin, Todor Mihaylov, Myle Ott, Sam Shleifer, Kurt Shuster, Daniel Simig, Punit~Singh Koura, Anjali Sridhar, Tianlu Wang, and Luke Zettlemoyer. 2022.
\newblock \href {http://arxiv.org/abs/2205.01068} {Opt: Open pre-trained transformer language models}.

\bibitem[{Zhong et~al.(2020{\natexlab{a}})Zhong, Yu, and Klein}]{zhong-etal-2020-semantic}
Ruiqi Zhong, Tao Yu, and Dan Klein. 2020{\natexlab{a}}.
\newblock \href {https://doi.org/10.18653/v1/2020.emnlp-main.29} {Semantic evaluation for text-to-{SQL} with distilled test suites}.
\newblock In \emph{Proceedings of the 2020 Conference on Empirical Methods in Natural Language Processing (EMNLP)}, pages 396--411, Online. Association for Computational Linguistics.

\bibitem[{Zhong et~al.(2020{\natexlab{b}})Zhong, Lewis, Wang, and Zettlemoyer}]{Zhong-etal-2020-grounded}
Victor Zhong, Mike Lewis, Sida~I. Wang, and Luke Zettlemoyer. 2020{\natexlab{b}}.
\newblock \href {https://doi.org/10.18653/v1/2020.emnlp-main.558} {Grounded adaptation for zero-shot executable semantic parsing}.
\newblock In \emph{Proceedings of the 2020 Conference on Empirical Methods in Natural Language Processing (EMNLP)}, pages 6869--6882, Online. Association for Computational Linguistics.

\bibitem[{Zhong et~al.(2017)Zhong, Xiong, and Socher}]{zhongSeq2SQL2017}
Victor Zhong, Caiming Xiong, and Richard Socher. 2017.
\newblock \href {http://arxiv.org/abs/1709.00103} {{Seq2SQL}: Generating structured queries from natural language using reinforcement learning}.

\end{thebibliography}

 \newpage
\clearpage
\appendix
\section{Data Generation}\label{app:data_generation}
Table~\ref{tab:full_prompts} shows  the full versions of the prompts we use to generate the  augmentations defined in Section~\ref{sec:data_generation} for the Spider training set.
\begin{table}[h!]
\centering
\footnotesize
\newcommand{\Item}{\stepcounter{Number}\theNumber.~}
\settowidth\tymin{\textbf{6. Instruction:}}
\begin{tabulary}{\linewidth}{@{}L@{\;}L@{}}\setcounter{Number}{0}
\textbf{\Item Instruction:} & \textbf{\texttt{Simplify}} \\ \addlinespace[3pt]
Full version & Simplify the following sentence: $\dots$ \\\midrule
\textbf{\Item Instruction:} & \textbf{\texttt{Simplify by hiding details}} \\ \addlinespace[3pt]
Full version & Simplify the sentence by hiding unnecessary details that do not change the meaning: $\dots$ \\\midrule
\textbf{\Item Instruction:} & \textbf{\texttt{Simplify using synonyms}} \\ \addlinespace[3pt]
Full version & Simplify the following sentence using synonyms: $\dots$ \\\midrule
\textbf{\Item Instruction:} & \textbf{\texttt{Simplify using substitutions}} \\ \addlinespace[3pt]
Full version & Make the sentence simpler by  substituting some words in $\dots$ \\\midrule
\textbf{\Item Instruction:} &   \textbf{\texttt{Express in a different way}} \\ \addlinespace[3pt]
Full version & What are different ways of expressing this question: $\dots$   \\\midrule
   \textbf{\Item Instruction:} & \textbf{\texttt{Examples of the question simplification: <$\dots$>}
   }\\ \addlinespace[3pt]
Full version & Examples of the question simplification:\\ \addlinespace[3pt]
& Original: Find the names of stadiums whose capacity is smaller than the average capacity.\\ \addlinespace[3pt]
&  Simplified: Which stadiums are smaller than the average?\\ \addlinespace[3pt]
&  Original: Show the fleet series of aircraft flown by pilots younger than 34.\\ \addlinespace[3pt]
&  Simplified: Return the fleet series of the planes whose captains are younger than 34.\\ \addlinespace[3pt]
& Original: Which cities have the largest population?\\ \addlinespace[3pt]
& Simplified: Where do most people live?\\ \addlinespace[3pt]
& Original: In which year was most of the ships built?\\ \addlinespace[3pt]
& Simplified: When were most of the ships constructed?\\ \addlinespace[3pt]
& Original: Tell me the number of orders with "Second time" as the order detail.\\ \addlinespace[3pt]
& Simplified: How many orders have "Second time" as an order detail?\\ \addlinespace[3pt]
& Original: $\dots$ \\ \addlinespace[3pt]
& Simplified: \\ \midrule
\textbf{\Item Instruction:} & \textbf{\texttt{Paraphrase}} \\ \addlinespace[3pt]
Full version &
Give me a paraphrase of the following question: $\dots$ 
\\ \bottomrule
\end{tabulary}
\caption{Prompts used for data generation.}
\label{tab:full_prompts}
\end{table}

\begin{table*}[t]
\centering
\footnotesize
\begin{tabular}{@{}l@{\;\;}l@{\;\;}l@{\;\;}ll@{\;\;}ll@{\;\;}ll@{\;\;}ll@{\;\;}ll@{\;\;}ll@{\;\;}l@{}}
\toprule
 & & \multicolumn{2}{c}{} & \multicolumn{2}{c}{Augmented}&\multicolumn{2}{c}{} & \multicolumn{2}{c}{Augmented} &\multicolumn{2}{c}{} & \multicolumn{2}{c}{Augmented} & \multicolumn{2}{c}{} \\\addlinespace[3pt]
 & & \multicolumn{2}{c}{T5-3B} & \multicolumn{2}{c}{T5-3B} & \multicolumn{2}{c}{PICARD} & \multicolumn{2}{c}{PICARD} & \multicolumn{2}{c}{RESDSQL} & \multicolumn{2}{c}{RESDSQL}  &   \multicolumn{2}{c}{ChatGPT}
 \\\addlinespace[3pt]
 \cmidrule(lr){3-4}\cmidrule(lr){5-6}\cmidrule(lr){7-8}\cmidrule(lr){9-10}\cmidrule(lr){11-12}\cmidrule(lr){13-14}\cmidrule(lr){15-16}
\multicolumn{2}{l}{Perturbation Set} & Pre & Post  & Pre & Post  & Pre & Post  & Pre & Post   & Pre & Post & Pre & Post  & Pre & Post   \\ \midrule
\multirow{10}{*}{NLQ} & Keyword-synonym & 70.2	& 62.6
& 73.8 & 65.4
& 72.6 & 66.3
& 75.3 & 69.4
& 81.5 & 72.4
& \textbf{84.2} & \textbf{74.7}
& 64.7 & 55.7
\\\addlinespace[3pt]
& Keyword-carrier & 82.7	& 76.4	
& 83.0 &	79.2 
& 85.0 & 82.7
&88.7 & 84.0
&  \textbf{89.0} & 83.5
& 87.5 &  \textbf{85.0}
& 85.0 &	82.0
\\\addlinespace[3pt]
& Column-synonym & 63.9 & 51.3 
& 66.3 & 54.2  
& 71.0 & 57.2
& 68.7 & \textbf{59.7}
& \textbf{78.7} & 63.1
& 77.4 & \textbf{66.1}
& 66.1 &	48.8 
\\\addlinespace[3pt]
& Column-carrier & 83.1 &	61.7	
& 82.0 & 70.5
& \textbf{86.9} & 64.9
& 85.0 & 73.1
& 86.5 & 63.9
& 86.4 & \textbf{76.3}
& 82.2 &	52.0 
\\\addlinespace[3pt]
& Column-attribute & 49.6 &	48.7 
& 60.5 &	58.8 
& 58.8  & 56.3
& 63.9 &	62.2
& \textbf{82.4} &	\textbf{71.4}
& \textbf{82.4} & \textbf{71.4}
& 77.3 & 62.2 
\\\addlinespace[3pt]
& Column-value & 69.1	& 58.6
& 76.3 & 58.9
& 82.9 & 69.4
& \textbf{83.2} & \textbf{70.4}
& \textbf{96.4} & 76.6
& 95.1 &	\textbf{77.6}
& 74.0 &	57.9 
\\\addlinespace[3pt]
& Value-synonym & 68.6 &	46.4 
& 68.6 & 53.0 
& 72.5 & 53.0
& 70.8 & \textbf{57.1}
& 79.2 & 53.2
& \textbf{79.6} & 55.1
& 69.0 &	45.8 
\\ \addlinespace[3pt]
& Multitype & 70.1	&51.1 
& 71.4 & 56.3 
& 74.4 & 57.1
& 74.0 & 61.4
& \textbf{83.8} & 60.7
& \textbf{83.8} & \textbf{65.7}
& 71.9 &	49.8 
\\\addlinespace[3pt]
& Others & 75.3	& 73.1	
& 76.6 &	72.7 
& 79.6 & \textbf{78.3}
& \textbf{80.9} & 77.6
& \textbf{85.2} &	79.0
& 84.8 & \textbf{80.2}
& 74.0	& 66.4 
\\\midrule
& Average & 70.3	&58.9
& 73.2 & 63.2
& 76.0 & 65.0
& 76.7 & 68.3
& \textbf{84.7} & 69.3
& 84.6 & \textbf{72.5}
& 73.8	& 57.9 \\\midrule
\multirow{6}{*}{SQL} & Comparison & 62.9& 	62.4& 
71.3 & 66.3 & 
68.0 &	68.0 &
74.2 &	70.8
& 80.9 & 82.0
& \textbf{84.3} & \textbf{83.7} &
73.6 & 64.0
\\\addlinespace[3pt]
& Sort-order  &
75.0&70.3 &
76.0 &	75.5 &
79.2 &	74.5 &
78.1 & 76.6
& 88.0 & \textbf{85.4}
& \textbf{88.5} & 83.3 &
66.7 &	57.8 \\\addlinespace[3pt]
& NonDB-number  &
77.1	& 73.3 &
71.8 &	77.1 &
83.2  & 77.1 &
73.3 & 77.9
& 87.8 & 85.5 &
 \textbf{90.8} & \textbf{90.8} &
\textbf{90.8} & 90.1
\\\addlinespace[3pt]
& DB-text  &
59.5	& 58.3 &
59.9 & 61.6 &
64.7& 65.1 & 
66.2 & 66.7
& 77.2 & 74.3%
& \textbf{91.5} & \textbf{75.0} &
67.5 & 68.2 \\\addlinespace[3pt]
& DB-number &
83.9 & 83.7 & 
79.8 & 78.8 & 
86.3 & 85.1 & 
84.6 &	83.2 
& 88.8 & 88.8
& \textbf{91.5} & \textbf{91.2}
& 82.7 &	79.8
\\\midrule
& Average & 71.7	& 69.6 & 71.8 & 71.9 & 76.3	& 74.0 & 75.3 &	75.0 &
84.5 &	83.2
& \textbf{89.3} & \textbf{84.8} &
76.3 &	72.0 
\\\midrule
\multirow{4}{*}{DB} & Schema-synonym & 
66.4& 46.9 &
67.8 & 52.8 &
73.0	& 56.5 &
73.4 & 61.9 &
 \textbf{81.3} &	68.3%
& 80.9 &  \textbf{70.4} &
67.6 & 56.0
\\\addlinespace[3pt]
& Schema-abbreviation &
69.5 & 53.3 & 
71.0 & 55.5 &
74.9 & 64.7 &
75.2 & 65.3 &
\textbf{82.4} &	70.0 &	81.8 & \textbf{71.7} &
68.8 &	63.5
\\\addlinespace[3pt]
& Content-equivalence &
84.6 & 40.8 & 
72.3 & 46.1 & 
88.7 & 43.7 &
86.9 & 37.2 & 
90.3 & 40.1%
&\textbf{91.9} &	41.4 &
81.2 & \textbf{46.3}
\\\midrule
& Average & 73.5 & 47.0 &72.3 & 46.1 & 78.9	& 55.0 & 78.5 & 54.8 
& 84.7 &	59.5
& \textbf{84.9} & \textbf{61.1}
& 72.5 & 55.3
\\ \midrule
\multicolumn{2}{l}{All} & 71.3 & 59.9 &72.6 & 62.7 &  76.6	& 65.9 &  76.6 & 67.9
& 84.7 &	71.7
& \textbf{86.0} &	\textbf{74.1}
&  74.3 &	61.5
\\ \bottomrule
\end{tabular}
\caption{Execution Accuracy on subsets  taken from Dr.Spider (NLQ, DB, and SQL sets); model performance is shown before  (Pre) and after perturbations (Post). We compare T5-3B, T5-3B+PICARD, and RESDSQL fine-tuned with and without augmentations, and zero-shot ChatGPT.}
\label{tab:robustness_full}
\end{table*}
\begin{table*}[!h]
\centering
\footnotesize
\begin{tabular}{lccccccc}
\toprule
 &  & Augmented &  & Augmented  &  & Augmented & \\
Dataset & T5-3B & T5-3B & PICARD & PICARD & RESDSQL & RESDSQL & ChatGPT \\\midrule
Realistic & 64.2 & 66.7 & 71.4 & 79.3 & 80.7 & \textbf{84.0} &  63.4 \\\addlinespace[3pt]
Spider-Syn & 62.4 &  70.8 & 69.8 & 72.8 & 76.9 & \textbf{79.2} & 58.6\\\addlinespace[3pt]
GeoQuery dev & 59.1 & 64.2 & 64.2 & \textbf{68.6} &  59.7 & 54.1  & 25.8 \\\bottomrule
\end{tabular}
\caption{Execution accuracy on  Spider-Realistic, Spider-Syn and GeoQuery dev set for T5-3B with and without PICARD and RESDSQL trained with or without augmentations. 
}
\label{tab:syn_real_geo}
\end{table*}

\section{Error Analysis}\label{app:error_analysis}
In order to verify that our augmentations do not introduce new parsing errors, we examined examples in the Spider development set  which were correctly parsed by a T5 model trained without augmentations but rendered incorrect after the same T5 model was trained with augmentations. Based on a sample of~60 instances, we observed that the majority of  errors are similar in nature and symptomatic of a T5-trained semantic parser, e.g., errors in the output columns or join operation. 

The only type of error that might be due to our augmentations concerns minor changes in values. Baseline T5 almost always copies values from the question but T5 trained with augmentations can slightly change them, e.g., use the full name instead of an abbreviation or lowercase instead of uppercase. We found this occurs in 10\% of cases. Database values are mentioned verbatim in Spider questions but this could be different in  real-world settings or other datasets where some tolerance to surface variations might be advantageous.

\section{ChatGPT Zero-Shot Prompt}
\label{app:chat_prompt}
Below we show the prompt we used when evaluating the zero-shot ChatGPT on text-to-SQL datasets following \citet{liu2023comprehensive}: 


\begin{lstlisting}[basicstyle=\footnotesize\ttfamily, breaklines=true]
### SQL tables, with their properties:
#
# stadium(Stadium_ID, Location, Name, Capacity, Highest, Lowest, Average)
# singer(Singer_ID, Name, Country, Song_Name, Song_release_year, Age, Is_male)
# concert(concert_ID, concert_Name, Theme, Stadium_ID, Year)
# singer_in_concert(concert_ID, Singer_ID)
#
### How many singers do we have? Return only a SQL query.
SELECT
\end{lstlisting}

\section{Additional Results}\label{app:add_results}

Table~\ref{tab:robustness_full} shows our results on \emph{all} Dr.Spider
perturbation subsets (NLQ refers to subsets with perturbations in natural language questions, SQL and DB are  perturbations in SQL and database tokens).
We compare three models trained with and without augmentations:
T5-3B, PICARD,  and RESDSQL. We also employ ChatGPT in a zero-shot setting.  Overall, the best model is augmented RESDSQL (74.1\%) which is better than the base version by more than 2\% on post-perturbed sets. Augmented T5-3B and PICARD also improve  robustness compared to  base models. Augmented RESDSQL  delivers the best average results for all three types of perturbations and performs best on the majority of individual categories, even though our augmentations are \emph{not} designed
to improve robustness against SQL and DB perturbations.

Table~\ref{tab:syn_real_geo} shows  results on the additional evaluation sets, Spider-Realistic, \citep{gan-etal-2021-towards} Spider-Syn with 1,034 examples, and GeoQuery dev set with 152 examples (query splits of \citealt{finegan-dollak-etal-2018-improving}). Both evaluation sets are based on the Spider development set,  aiming to remove from the questions explicit references to database entities. These references were manually deleted or paraphrased in Spider-Realistic  and replaced with synonyms in Spider-Syn. Augmented RESDSQL obtains best results on both datasets (84.0\% on Spider-Realistic and 79.2\% on Spider-Syn) and is better than the base version by more than~4\%. On the GeoQuery development set, the best model is augmented PICARD with~68.6\% accuracy. Across \emph{all} benchmarks, fine-tuned text-to-SQL parsers significantly outperform zero-shot ChatGPT.

\section{Examples of Spider Augmentations}
\label{sec:examples:augmented}
We provide samples of the augmented Spider training set. Questions are grouped based on intent; types indicate whether they are in the \textit{original} Spider training set  or  were generated with  instructions: \textit{simplify},  simplify by \textit{hiding details},  simplify by \textit{synonyms},  simplify by \textit{substitutions},  \textit{express differently},  \textit{paraphrase}, or by showing \textit{examples}.

\begin{table*}
\centering
\footnotesize
\newcommand{\Item}{\stepcounter{Number}\theNumber.~}
\settowidth\tymin{\textbf{Type: express-differently}}
\begin{tabulary}{\textwidth}{@{}L@{\;}L@{\;}LL@{}}\setcounter{Number}{0}
\Item & SQL query & \multicolumn{2}{l}{\texttt{SELECT personal\_name FROM Students EXCEPT SELECT T1.personal\_name FROM  }} \\\addlinespace[1pt]
& & \multicolumn{2}{l}{\texttt{Students AS T1 JOIN Student\_Course\_Enrolment AS T2 ON T1.student\_id  =  T2.student\_id}} \\\addlinespace[3pt]
& Questions & Find the personal names of students not enrolled in any course. & Type: original \\\addlinespace[3pt]
& & Which students not enrolled in any course? Find their personal names.  & Type: original \\\addlinespace[3pt]
& & Find names of unenrolled students. & Type: hide-details \\\addlinespace[3pt]
& &  Can you provide me with the names of the students who are not currently registered in any course? & Type: paraphrase\\\midrule
\Item & SQL query & \multicolumn{2}{l}{\texttt{SELECT donator\_name ,  sum(amount) FROM endowment GROUP BY donator\_name}}  \\\addlinespace[1pt]
& & \multicolumn{2}{l}{\texttt{ORDER BY sum(amount) DESC}} \\\addlinespace[3pt]
& Questions & List each donator name and the amount of endowment in descending order of the amount of endowment & Type: original \\\addlinespace[3pt]
& & Enumerate contributors and their endowment sums in decreasing order & Type: synonyms\\\addlinespace[3pt]
& & List donors and their endowments in descending order & Type: hide-details\\\midrule
\Item & SQL query & \multicolumn{2}{l}{\texttt{SELECT count(*) FROM CLASS AS T1 JOIN enroll AS T2 ON T1.class\_code  =  T2.class\_code }}  \\\addlinespace[1pt]
 & & \multicolumn{2}{l}{\texttt{WHERE T1.crs\_code  =  'ACCT-211'}}  \\\addlinespace[3pt]
 & Questions & How many students enrolled in class ACCT-211? & Type: original \\\addlinespace[3pt]
  &  & What are the total number of students enrolled in ACCT-211? & Type: original \\\addlinespace[3pt]
  & & How many pupils registered for course ACCT-211? & Type: synonyms \\\addlinespace[3pt]
 & & How many students are enrolled in ACCT-211? & Type: from-examples \\\midrule
\Item & SQL query & \multicolumn{2}{l}{\texttt{SELECT T2.roomName FROM Reservations AS T1 JOIN Rooms AS T2 ON T1.Room  =  T2.RoomId  }}  \\\addlinespace[1pt]
 &  & \multicolumn{2}{l}{\texttt{WHERE firstname LIKE '\%ROY\%'}}  \\\addlinespace[3pt]
& Questions & Find the name of rooms booked by some customers whose first name contains ROY. & Type: original \\\addlinespace[3pt]
& & What are the name of rooms booked by customers whose first name has "ROY" in part? & Type: original \\\addlinespace[3pt]
& & What are the room names that have been reserved by customers with "ROY" in their first name? & Type: paraphrase \\\addlinespace[3pt]
& & What rooms did customers with "ROY" in their first name book?& Type: substitutions \\\midrule
\Item & SQL query & \multicolumn{2}{l}{\texttt{SELECT T1.account\_name ,  T1.other\_account\_details FROM Accounts AS T1 JOIN }}  \\\addlinespace[1pt]
 &  & \multicolumn{2}{l}{\texttt{Customers AS T2 ON T1.customer\_id  =  T2.customer\_id WHERE  }}  \\\addlinespace[1pt]
  &  & \multicolumn{2}{l}{\texttt{T2.customer\_first\_name  =  'Meaghan' AND T2.customer\_last\_name  =  'Keeling'}}  \\\addlinespace[3pt]
& Questions & Show the account name and other account detail for all accounts by the customer with first name Meaghan and last name Keeling. & Type: original \\\addlinespace[3pt]
& & What are the names and other details for accounts corresponding to the customer named Meaghan Keeling? & Type: original \\\addlinespace[3pt]
& & What are Meaghan Keeling's account names and details? & Type: simplify \\\addlinespace[3pt]
& & I am looking for the account details and names associated with Meaghan Keeling. Can you help me with that? & Type: paraphrase \\\midrule
\Item & SQL query & \multicolumn{2}{l}{\texttt{SELECT sum(acc\_bal) FROM customer WHERE state  =  'Utah' OR state  =  'Texas'}}   \\\addlinespace[3pt]
& Questions & Find the total account balance of each customer from Utah or Texas. & Type: original \\\addlinespace[3pt]
& & What are the total account balances for each customer from Utah or Texas?  & Type: original \\\addlinespace[3pt]
& & Add up the account balances of customers who live in Utah or Texas. & Type: express-differently \\\addlinespace[3pt]
& & What is the total account balance for customers from Utah or Texas? & Type: from-examples \\\midrule
\Item & SQL query & \multicolumn{2}{l}{\texttt{SELECT date\_of\_enrolment , date\_of\_completion FROM Student\_Course\_Enrolment}}  \\\addlinespace[3pt]
& Questions & List all the dates of enrollment and completion of students. & Type: original \\\addlinespace[3pt]
& & What are all the dates of enrollment and completion in record? & Type: original \\\addlinespace[3pt]
& & Provide a record of the enrollment and completion dates for all students. & Type: paraphrase \\\addlinespace[3pt]
& & What are the enrollment and completion dates of all students? & Type: from-examples\\\midrule
\Item & SQL query & \multicolumn{2}{l}{\texttt{SELECT headquarter FROM manufacturers WHERE founder  =  'James'}}  \\\addlinespace[3pt]
& Questions & Where is the headquarter of the company founded by James? & Type: original \\\addlinespace[3pt]
& & What is the headquarter of the company whose founder is James? & Type: original \\\addlinespace[3pt]
& & Where was the company founded by James headquartered? & Type: express-differently \\\addlinespace[3pt]
& & Where is the main office of the company established by James? & Type: paraphrase 
\end{tabulary}
\end{table*}
\newpage
\begin{table*}
\centering
\footnotesize
\newcommand{\Item}{\stepcounter{Number}\theNumber.~}
\settowidth\tymin{\textbf{Type: express-differently}}
\begin{tabulary}{\textwidth}{@{}L@{\;}L@{\;}LL@{}}\setcounter{Number}{8}
\Item & SQL query & \multicolumn{2}{l}{\texttt{SELECT max(Price) ,  max(Score) FROM WINE WHERE Appelation  =  'St. Helena'}}  \\\addlinespace[3pt]
& Questions & What are the maximum price and score of wines produced by St. Helena appelation? & Type: original \\\addlinespace[3pt]
& & Give the maximum price and score for wines produced in the appelation St. Helena. & Type: original \\\addlinespace[3pt]
& & What is the topmost price and score that can be obtained by wines produced in St. Helena? & Type: paraphrase \\\addlinespace[3pt]
& & What is the highest price and score for St. Helena wines?  & Type: simplify \\\midrule
\Item & SQL query & \multicolumn{2}{l}{\texttt{SELECT degrees FROM campuses AS T1 JOIN degrees AS T2 ON t1.id  =  t2.campus }} \\\addlinespace[1pt]
&  & \multicolumn{2}{l}{\texttt{WHERE t1.campus  =  'San Francisco State University' AND t2.year  =  2001}} \\\addlinespace[3pt]
& Questions &  What are the degrees conferred in "San Francisco State University" in 2001. & Type: original \\\addlinespace[3pt]
& & What degrees were conferred in San Francisco State University in the year 2001? & Type: original \\\addlinespace[3pt]
& & What diplomas were granted at SF State in 2001? & Type: synonyms \\\addlinespace[3pt]
& & What degrees were given at San Francisco State University in 2001? & Type: substitutions \\\midrule
\Item & SQL query & \multicolumn{2}{l}{\texttt{SELECT membership\_card FROM member WHERE address  =  'Hartford' INTERSECT }}  \\\addlinespace[1pt]
& & \multicolumn{2}{l}{\texttt{SELECT membership\_card FROM member WHERE address  =  'Waterbury'}} \\\addlinespace[3pt]
& Questions & What is the membership card held by both members living in Hartford and ones living in Waterbury address?  & Type: original \\\addlinespace[3pt]
& & What is the membership card for people in Hartford and Waterbury called?  & Type: substitutions \\\addlinespace[3pt]
& & Is there a membership card that is valid for both Hartford and Waterbury residents?  & Type: express-differently \\\midrule
\Item & SQL query & \multicolumn{2}{l}{\texttt{SELECT kids FROM Reservations WHERE FirstName = 'ROY' AND LastName  =  'SWEAZY'}}  \\\addlinespace[3pt]
& Questions & How many kids stay in the rooms reserved by ROY SWEAZY? & Type: original \\\addlinespace[3pt]
& & Find the number of kids staying in the rooms reserved by a person called ROY SWEAZ. & Type: original \\\addlinespace[3pt]
& & How many children are staying in ROY SWEAZY's reserved rooms? & Type: from-examples \\\addlinespace[3pt]
& & How many kids are in Roy Sweaz's reserved rooms? & Type: hide-details \\\midrule
\Item & SQL query & \multicolumn{2}{l}{\texttt{SELECT count(*) FROM products AS t1 JOIN product\_characteristics AS t2 }}  \\\addlinespace[1pt]
&  & \multicolumn{2}{l}{\texttt{ON t1.product\_id  =  t2.product\_id JOIN CHARACTERISTICS AS t3 }} \\\addlinespace[1pt]
&  & \multicolumn{2}{l}{\texttt{ON t2.characteristic\_id  =  t3.characteristic\_id WHERE t1.product\_name  =  'laurel'}}\\\addlinespace[3pt]
& Questions & How many characteristics does the product named "laurel" have?  & Type: original \\\addlinespace[3pt]
&  & Count the number of characteristics of the product named 'laurel'. & Type: original \\\addlinespace[3pt]
& & How many features does "laurel" have?  & Type: simplify \\\addlinespace[3pt]
& & How many qualities does the product "laurel" have? & Type: substitutions \\\midrule
\Item & SQL query & \multicolumn{2}{l}{\texttt{SELECT customer\_name FROM customers WHERE payment\_method = (SELECT payment\_method }}  \\\addlinespace[1pt]
&  & \multicolumn{2}{l}{\texttt{FROM customers GROUP BY payment\_method ORDER BY count(*) DESC LIMIT 1)}} \\\addlinespace[3pt]
& Questions & What are the names of customers using the most popular payment method? & Type: original \\\addlinespace[3pt]
& & Find the name of the customers who use the most frequently used payment method. & Type: original \\\addlinespace[3pt]
& & Who are the customers using the popular payment method?  & Type: hide-details  \\\addlinespace[3pt]
& & Who are the customers utilizing the most favored payment option? & Type: synonyms \\\midrule
\Item & SQL query & \multicolumn{2}{l}{\texttt{SELECT TYPE FROM ship WHERE Tonnage  >  6000 INTERSECT SELECT TYPE FROM ship }} \\\addlinespace[1pt]
&  & \multicolumn{2}{l}{\texttt{WHERE Tonnage  <  4000}} \\\addlinespace[3pt]
& Questions & Show the types of ships that have both ships with tonnage larger than 6000 and ships with tonnage smaller than 4000. & Type: original \\\addlinespace[3pt]
& & What are the types of the ships that have both shiips with tonnage more than 6000 and those with tonnage less than 4000? & Type: original \\\addlinespace[3pt]
& & Display ships with tonnage above 6000 and below 4000. & Type: simplify \\\addlinespace[3pt]
& & Which types of ships have tonnage exceeding 6000 and also less than 4000? & Type: express-differently \\\midrule
\Item & SQL query & \multicolumn{2}{l}{\texttt{SELECT customer\_name FROM customers EXCEPT SELECT t1.customer\_name FROM customers AS t1 }} \\\addlinespace[1pt]
&  & \multicolumn{2}{l}{\texttt{JOIN customer\_addresses AS t2 ON t1.customer\_id  =  t2.customer\_id JOIN addresses AS t3}}\\\addlinespace[1pt]
&  & \multicolumn{2}{l}{\texttt{ON t2.address\_id  =  t3.address\_id WHERE t3.state\_province\_county  =  'California'}} \\\addlinespace[3pt]
& Questions & Find the names of customers who are not living in the state of California & Type: original \\\addlinespace[3pt]
& & Discover the names of non-California customers. & Type: substitutions \\\addlinespace[3pt]
& & Who are the customers not residing in California? & Type: from-examples \\\addlinespace[3pt]
\end{tabulary}
\end{table*}

\end{document}